\documentclass[letterpaper]{article} 
\usepackage{aaai25}  
\usepackage{times}  
\usepackage{helvet}  
\usepackage{courier}  
\usepackage[hyphens]{url}  
\usepackage{graphicx} 
\usepackage{natbib}  
\usepackage{caption} 
\usepackage[linesnumbered,ruled,vlined]{algorithm2e}
\usepackage[table]{xcolor}
\usepackage{xspace}
\usepackage{amsmath}
\usepackage{amsfonts}
\usepackage{cleveref}
\usepackage{booktabs}
\usepackage{multirow}
\usepackage{newfloat}

\title{\textsc{Latte}: Improving \underline{LaT}eX Recognition with I\underline{t}erative R\underline{e}finement}
\author {
    Nan Jiang\textsuperscript{\rm 1},
    Shanchao Liang\textsuperscript{\rm 1},
    Chengxiao Wang\textsuperscript{\rm 1, \rm 2},
    Jiannan Wang\textsuperscript{\rm 1},
    Lin Tan\textsuperscript{\rm 1}
}
\affiliations {
    \textsuperscript{\rm 1} Purdue University, USA\\
    \textsuperscript{\rm 2} University of Illinois Urbana-Champaign, USA \\
    jiang719@purdue.edu, liang422@purdue.edu, cw124@illinois.edu, wang4524@purdue.edu, lintan@purdue.edu
}

\usepackage{bibentry}

\begin{document}

\maketitle

\definecolor{lightgreen}{RGB}{216,255,216} 
\definecolor{lightblue}{RGB}{216,216,255} 
\definecolor{lightred}{RGB}{255,216,216} 

\newcommand{\ours}{\textsc{Latte}\xspace}
\newcommand{\oursone}{\textsc{Latte$_1$}\xspace}
\newcommand{\ourstwo}{\textsc{Latte$_2$}\xspace}
\newcommand{\oursthree}{\textsc{Latte$_3$}\xspace}
\newcommand{\oursfour}{\textsc{Latte$_4$}\xspace}

\newcommand{\ete}{\textsc{Latte$_{-fl-dv}$}\xspace}
\newcommand{\etedv}{\textsc{Latte$_{-fl}$}\xspace}

\newcommand{\idiff}{{\code{ImageEdit}}\xspace}
\newcommand{\latex}{{\small LaTeX}\xspace}
\newcommand{\deltaimage}{\code{delta-view}\xspace}
\newcommand{\Deltaimage}{\code{Delta-View}\xspace}
\newcommand{\pairedimage}{\code{paired-view}\xspace}
\newcommand{\Pairedimage}{\code{Paired-View}\xspace}
\newcommand{\tabletolatex}{{\small T{\scriptsize AB}2L{\scriptsize ATEX}}\xspace}
\newcommand{\imgtolatex}{{\small I{\scriptsize MG}2L{\scriptsize ATEX}-100{\scriptsize K}}\xspace}
\newcommand{\faultloc}{{fault localization model}\xspace}
\newcommand{\Faultloc}{{Fault Localization Model}\xspace}
\newcommand{\code}[1]{\texttt{\small #1}}

\newcommand{\ie}{i.e.\xspace}
\newcommand{\eg}{e.g.\xspace}

\definecolor{lightgray}{gray}{0.85}

\captionsetup[figure]{skip=8pt}
\captionsetup[table]{skip=8pt}
\newcommand{\distance}{10pt}
\setlength{\textfloatsep}{\distance}
\setlength{\floatsep}{\distance}
\setlength{\intextsep}{\distance}
\setlength{\dbltextfloatsep}{\distance} 
\setlength{\dblfloatsep}{\distance} 

\begin{abstract}
Portable Document Format (PDF) files are dominantly used for storing and disseminating scientific research, legal documents, and tax information. \latex{} is a popular application for creating PDF documents. Despite its advantages, \latex{} is not WYSWYG---what you see is what you get, i.e.,  the \latex{} source and rendered PDF images look drastically different, especially for formulae and tables. This gap makes it hard to modify or export \latex{} sources for formulae and tables from PDF images, and existing work is still limited. First, prior work generates \latex{} sources in a single iteration and struggles with complex \latex{} formulae. Second, existing work mainly recognizes and extracts \latex{} sources for formulae; and is incapable or ineffective for tables. This paper proposes \ours{}, the first \emph{iterative refinement} framework for \latex{} recognition. Specifically, we propose \deltaimage as feedback, which compares and pinpoints the differences between a pair of rendered images of the extracted \latex{} source and the expected correct image. Such \deltaimage feedback enables our fault localization model to localize the faulty parts of the incorrect recognition more accurately and enables our \latex{} refinement model to repair the incorrect extraction more accurately. \ours{} improves the \latex{} source extraction accuracy of both \latex{} formulae and tables, outperforming existing techniques as well as GPT-4V by at least 7.03\% of exact match, with a success refinement rate of 46.08\% (formula) and 25.51\% (table).
\end{abstract}

\begin{figure*}[htp]
    \centering
    \includegraphics[width=0.9\textwidth]{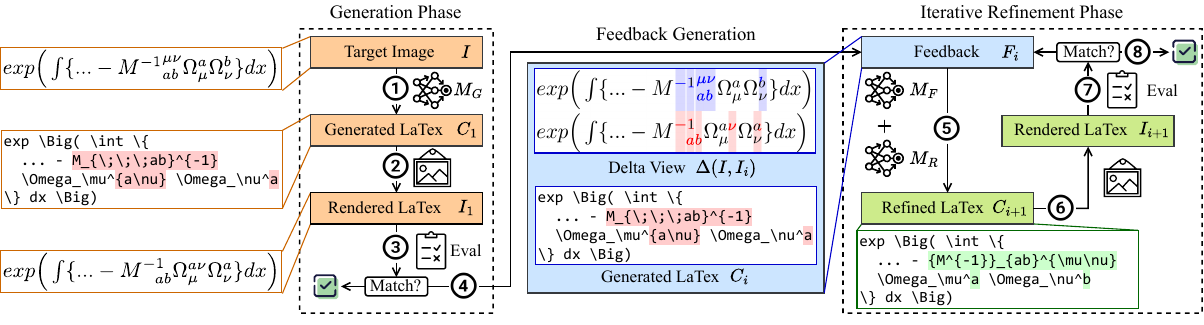}
    \caption{Overview of \ours. $M_G$ is the initial \latex source generation model, $M_F$ is the fault localization model, and $M_R$ is the refinement model.}
    \label{fig:overview}
\end{figure*}

\section{Introduction}
Portable Document Format (PDF) files are dominantly used for storing and disseminating academic research, legal documents, and tax information~\cite{nougat,pdf-related1}. While documents in such format provide exceptional cross-platform consistency and readability and are flexible in display resolutions, the source code of the PDF files is typically unavailable to the readers. Thus,  it is hard to modify, extract, and export PDF documents to other target formats, especially those containing mathematical formulae and tables with complex structures and styles.

Since \latex is one widely used system to produce PDF documents~\cite{xpdf}, researchers have explored \latex recognition (LR)  to extract mathematical expressions from images using either rule-based or learning-based approaches~\cite{related1, related2, im2latex, related3, related4, related5, related6}. Other work focuses on extracting table structures or detecting the content in each cell~\cite{table2latex-related2-overview, table2latex-related4-seperate}, which helps understand and analyze tables. 

These existing techniques~\cite{related1, related2, im2latex, related3, related4, related5, related6} is limited in recognizing \latex images. First, they produce \latex sources in a single round of generation and often fail to recognize complex formulae.
Our insight is that \emph{humans often write complex formulae and tables in multiple iterations}. For example, if the first version of the \latex source is incorrect, they fix the mistakes, re-render the modified \latex source, and keep this iterative process.
Second, existing techniques focus on \latex formulae. The few table recognition techniques do not extract ready-to-use \latex source for tables~\cite{table2latex-related2-overview, table2latex-related4-seperate}, but only extract table structures or textual content. Simply combining the table structures and content does not produce  \latex sources that can be rendered~\cite{table2latex-related4-seperate},  because the structures and content may mismatch with each other.

In this paper, we propose \textbf{\ours{}}, a \latex recognition framework with \textbf{iterative refinement}. 
We also create a new \latex table dataset, \tabletolatex, by collecting \latex tables source code from arXiv preprints and the corresponding \latex sources. \tabletolatex is a dataset for end-to-end \latex tables recognition, aiding the development of techniques to produce renderable \latex sources for tables.
We demonstrate the effectiveness of \ours on recognizing \textbf{both \latex tables and formulae}.

The concept of iterative refinement has been applied in various fields, including code generation and code refinement~\cite{llm-selfrefine-related1, llm-selfrefine-related2, llm-selfrefine-related3, llm-selfrefine-related4, llm-selfrefine-related5}. The process typically consists of two parts: generating an initial draft and then iteratively refining it using collected feedback until it meets the requirements.
Yet, applying the refinement framework in the field of LR is challenging, as it is hard to generate feedback that effectively connects the expected ground-truth image and the generated textual draft. The difference between the expected image and the image rendered from the draft needs to be identified automatically. And the model is also required to learn the portions of the generated text that cause such differences in images, \ie, building the connection between textual scripts and rendered images. To overcome this challenge, we propose an \idiff algorithm to pinpoint the differences between the ground truth and rendered images, referred to as \deltaimage. \textbf{We use delta-view as feedback} to aid \ours to localize and refine the error.

Another challenge of applying the refinement framework in LR is identifying the faulty location of the generated \latex script. The faulty \latex scripts typically are only incorrect in a small portion, such as a few incorrect characters for mathematical formulae, or a few cells for tables. Instead of re-generating the whole \latex script, one can localize the faulty parts and re-generate those parts only. Thus, we implement a \faultloc trained along with the refinement model to predict the faulty location. Once we surmount this challenge and successfully identify the faulty location of the \latex script, \ours{} only need to re-generate the incorrect portion of it, which minimizes the learning challenges of the refinement model.

To sum up, this paper makes the following contributions:
\begin{itemize}
    \item We create the first iterative-refinement approach for \latex recognition, which fine-tunes a localization model to identify the faulty part in \latex sources and use a refinement model to regenerate the faulty part of the \latex sources iteratively. 
    \item We propose a novel algorithm, \idiff, which produces effective feedback, \deltaimage, to enhance the refinement accuracy.
    \item We collect and open-source a new dataset for \latex tables recognition, \tabletolatex, filling the blank of no published dataset for end-to-end \latex table recognition.
    \item By combining iterative-refinement and \idiff, we develop \textbf{\ours{}} to produce renderable \latex code for both formulae and tables, outperforming existing techniques on formulae by 7.07\% of exact match, and commercial tools on tables by 56.00\%, with an overall fault localization accuracy of 56.90--60.53\%, and refinement rate of 25.51--46.08\%.
\end{itemize}

\section{Approach}
\Cref{fig:overview} provides an overview of \ours, which consists of two phases --- the Generation Phase and the Iterative Refinement Phase. 
Given the target document image $I$ to recognize, the generation model $M_G$ generates a \latex output $C_1$ as the initial draft (step \raisebox{.5pt}{\textcircled{\raisebox{-.9pt} {1}}}). \ours then uses \code{pdflatex} to render the \latex source draft  into an image $I_1$  (step \raisebox{.5pt}{\textcircled{\raisebox{-.9pt} {2}}}) and compares it with the ground-truth image $I$ (step \raisebox{.5pt}{\textcircled{\raisebox{-.9pt} {3}}}). If they match at the pixel level, signaling that the \latex source $C_1$ is correct, the process ends and \ours outputs $C_1$. 

Otherwise, \ours{} enters the Refinement Phase (step \raisebox{.5pt}{\textcircled{\raisebox{-.9pt} {4}}}).
During the $i^\text{th}$ iteration of the refinement phase, \ours{} automatically generates feedback $F_i$ consisting of $C_i$ and \deltaimage$\Delta(I, I_i)$, highlighting the difference between the ground truth and the rendered image. 
Then, \ours{} uses the \faultloc, $M_F$, to predict the faulty location in the \latex script. The predicted location is used to construct the input for the refinement model $M_R$. $M_R$ generates the refined \latex script starting from the predicted faulty location (step \raisebox{.5pt}{\textcircled{\raisebox{-.9pt} {5}}}), which replaces the faulty parts in $C_i$ to form the fully refined script $C_{i+1}$.
The refined script $C_{i+1}$ is rendered into a new image $I_{i+1}$ (step \raisebox{.5pt}{\textcircled{\raisebox{-.9pt} {6}}}), and is compared to the ground-truth image for evaluation (step \raisebox{.5pt}{\textcircled{\raisebox{-.9pt} {7}}}). Such a refinement phase continues until the evaluation passes (step \raisebox{.5pt}{\textcircled{\raisebox{-.9pt} {8}}}) or reaches the iteration limit.

\subsection{Generation Phase}
\ours{}'s generation model, $M_G$, is fine-tuned on top of the Nougat-base~\cite{nougat}, a multi-modal vision-encoder-decoder~\cite{vision-encoder-decoder} LLM pre-trained on RGB images of academic documents and their markdown sources. 
The input for the $M_G$ is an image $I \in \mathbb{N}^{H \times W \times 3}$ of a rendered formula or a table, where $H$ and $W$ represent the height and width of the image respectively, and $3$ refers to the color channels in RGB images. The vision-encoder of $M_G$ encodes the image, and the text decoder generates the corresponding \latex source code of the input image.

\subsection{Evaluation and Feedback Generation}
\label{sec:feedback}
After the generation model $M_G$ produces the initial \latex draft, \ours{} evaluates its correctness by rendering it using renderer \code{pdflatex}, and comparing it with the ground-truth image. If the rendered image matches the ground-truth image, the generated \latex script will be returned without refinement. Otherwise, it needs to be refined.

The feedback, which is the input to the refinement model, contains two elements: \deltaimage $\Delta(I, I_i)$ and the generated \latex script of the current refinement iteration $C_{i}$. To facilitate the localization of the fault in the incorrect script $C_i$, this work proposes the \idiff algorithm, which highlights the differences between the rendered image $I_i$ and the ground-truth image $I$, and generates $\Delta(I, I_i)$. The \idiff algorithm is based on the Wagner–Fischer algorithm~\cite{related-LDdistance-algo} used for computing the Levenshtein-Distance. \idiff treats \latex images as lists of columns of pixels. It calculates the least number of insertions, deletions, and substitutions of columns needed to transform the rendered image to the ground truth image, which is marked by light blue or light red backgrounds. Details can be found in the supplementary materials.

\begin{figure}[t]
    \centering
    \includegraphics[width=0.95\linewidth]{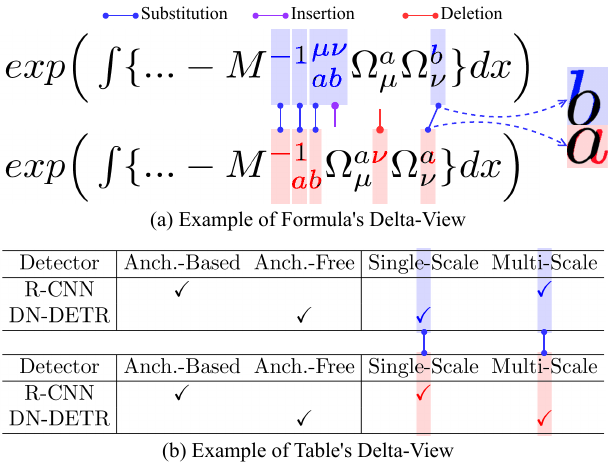}
    \caption{Formula and table examples of \deltaimage generated by the \idiff algorithm.}
    \label{fig:delta-view}
\end{figure}

\Cref{fig:delta-view} (a) provides an example of computing the \deltaimage for \latex formula. \idiff uses four blocks of substitutions, one block of deletion, and one block of insertion to show the difference. For example, the \raisebox{0pt}[0.8\height][0pt]{\colorbox{lightred}{\makebox(4,7){\textcolor{black}{$^a_\nu$}}}}
in the rendered image is incorrect and should be replaced by the \raisebox{0pt}[0.8\height][0pt]{\colorbox{lightblue}{\makebox(4,7){$^b_\nu$}}} in the ground truth.
In addition, \idiff provides finer-grained differences to help $M_F$ generate more accurate refined \latex sources. For the substitution from \raisebox{0pt}[0.8\height][0pt]{\colorbox{lightred}{\makebox(4,7){$^a_\nu$}}} to \raisebox{0pt}[0.8\height][0pt]{\colorbox{lightblue}{\makebox(4,7){$^b_\nu$}}}, \idiff identifies that only a portion of $a$ and $b$ is different, which is highlighted in blue and red. The identical parts, \ie, $\nu$ and a portion of $a$ and $b$, are shown in black.

For the table example shown in~\Cref{fig:delta-view} (b), the colored backgrounds mark the mismatched columns, while the blue and red edits show that the real fault is the positions of the checks in each column. In addition, to handle the more complex 2-D structure of table images, both the column-wised Levenshtein-Distance and the row-wised Levenshtein-Distance are calculated and the \deltaimage is generated using the solution with fewer edit percentages.

\subsection{Iterative-Refinement Phase}
The refinement phase of \ours{} consists of two steps: fault localization and refinement. The fault localization model pinpoints the faulty portion in the incorrect \latex script, which enables the refinement model to focus on modifying the wrong portion. The refinement model then generates the refined \latex script to replace the faulty portion suggested by the localization model.

\subsubsection{Fault Localization Model}
\begin{figure}[t]
    \centering
    \includegraphics[width=0.95\linewidth]{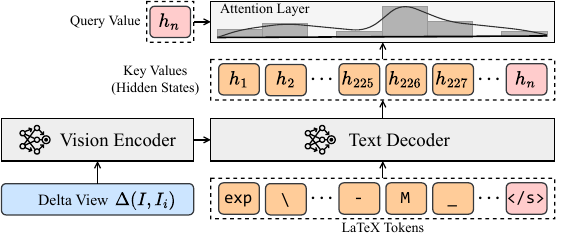}
    \caption{Fault localization model architecture.}
    \label{fig:fault_loc}
\end{figure}

\ours's fault localization model predicts the location of the first erroneous token in the corresponding \latex script. ~\Cref{fig:fault_loc} shows the \faultloc's architecture. 

The \faultloc includes a vision-encoder-decoder (VED) model and an attention layer. Given the incorrect \latex script $C_i = \{c_1, \ldots, c_n\}$ and the \deltaimage $\Delta (I, I_i)$ as input, the vision-encoder-decoder model calculates the hidden states of $c_i$ to $c_n$ as shown in \Cref{eq:fl} (notated by $H$).

The following attention layer calculates the attention scores on each token $c_i$, using its hidden states $H$ as keys and $h_n$ as the query. The reason for using $h_n$, the hidden states of the last token \code{</s>} at the end of the incorrect \latex script, as the query for calculating attention score is that: $h_n$ is the only hidden states in $H$ that incorporates the features of the entire $C_i$. As in the text decoder, every other token only incorporates the features of tokens before them, missing the global view of the whole incorrect \latex script.

In the attention layer, $W_q, W_k$ are trainable weights to encode the query and keys to compute the attention score distribution $P$.
Once $P$ is obtained, the index with the highest attention score will be selected as the faulty location $l$. The full formulation of fault localization is as follow:

\begin{equation}
\small
\begin{aligned}
    & H = \text{VED}(C_i, \Delta (I, I_i)) \\
    & Q = \text{ReLU}(W_q \cdot h_n) \:\;\;, \;\: K = \text{ReLU}(W_k \cdot H) \\
    & P = \text{Softmax}\left(Q K^{\top}\right), \;\; l = \underset{1 \leq i \leq n}{\mathrm{argmax}}(P)
\end{aligned}
\label{eq:fl}
\end{equation}

The training objective for the \faultloc is to minimize the Negative Log-Likelihood (NLL) loss on the probability of predicting the ground-truth faulty location $l_i$ for the given \latex script $C_i$ to refine by updating the fault localization model's weights $\theta_{M_F}$.

\subsubsection{Refinement Model with Fault Location}
\begin{figure}[!t]
    \centering
    \includegraphics[width=\linewidth]{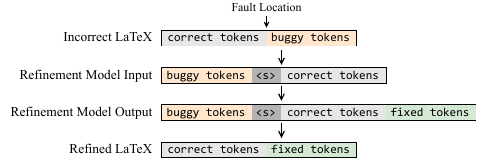}
    \caption{Workflow of the refinement model.}
    \label{fig:refine_prompt}
\end{figure}

As~\Cref{fig:refine_prompt} shows, given a wrong \latex $C_i$ to refine and the faulty location $l_i$ of it, the textual input for refinement model is structured as follows: ``\code{$C_i[l_i:]$ <s> $C_i[:l_i]$}''. 

Different from inputting the whole incorrect script $C_i$ as input and training the refinement model to generate the whole refined script, this template utilizes the faulty location by putting the faulty part of $C_i$ at the beginning of the prompt. \code{<s>} is used as a separator and tokens after it are correct parts (tokens with light grey background). Such an input format design is more effective than inputting the whole incorrect script as is~\cite{deepdive-apr}.
The refinement model is fine-tuned to generate the refined \latex tokens replacing the faulty parts (\eg, generating the \latex script with green background in~\Cref{fig:refine_prompt}). The final refined \latex script can be easily reconstructed from the prompt and refinement model's generation, which is the non-faulty parts (\latex script before the faulty location) followed by the refinement model's generation.

Formally, given the incorrect \latex script $C_i$ and faulty location $l_i$, we notate the ground-truth of the refined part to be $R_i = \{r_1, r_2, \ldots, r_m\}$, then the training objective of the refinement model is minimizing the negative log-likelihood of generating the $R_i$ based on the prompt by updating the model's weights $\theta_{M_R}$:

\begin{equation}
\small
\begin{aligned}
    L_R(\theta_{M_R}) = -\log(P(R_i | \{&c_{l_i}, c_{l_i+1}, \ldots, c_n, \texttt{<\textbackslash s>}, \\
    & c_1, \ldots, c_{l_i-1}\}, \Delta (I,I_i)))
\end{aligned}
\end{equation}

During inference, the predicted faulty location $l'_i$ generated by the fault localization model is used to build the prompt for the refinement model. The refinement model will generate the refined part $R'_i = \{r'_1, r'_2, \ldots, r'_{m'}\}$. The final refined \latex script is constructed as $C_{i+1} = \{c_{1}, \ldots, c_{l'_i-1}, r'_1, \ldots, r'_{m'}\}$.

\section{Experimental Setup}

\subsection{Datasets}
For formulae recognition, we use \textbf{\imgtolatex}, which consists of 103,556 rendered images of mathematical formulae and the corresponding \latex scripts, collected from over 60,000 published academic documents~\cite{im2latex}. During preprocessing, the source \latex scripts are first rendered to PDF format and then converted to PNG format with 240 dpi. Then, the PNG images are either resized or padded to the resolution of $1344\times224$ pixels.

For table recognition, as there are no open-sourced datasets for end-to-end \latex table recognition yet, this work constructs a new dataset, \textbf{\tabletolatex}. 
\tabletolatex consists of 97,532 rendered images of tables (resolution of $1344\times672$ pixels) and their \latex sources. The collection of this dataset is in the supplementary materials.

\subsection{Formula Models Training}
We use the default split of the \imgtolatex dataset, which has 73,812 training, 18,672 validation, and 10,072 test instances, to train the generation model.
We fine-tune the pre-trained Nougat-base model~\cite{nougat}, using a batch size of 16. The model weights are optimized using the AdamW~\cite{adamw} optimizer, with the learning rate set to $3e^{-5}$, using 1,000 steps of warm-up and adjusted by a cosine decay scheduler.

For fault localization and refinement models, we collect incorrect \latex sources by sampling 20 \latex sources per image in the training set of \imgtolatex (sampling temperature set to 0.8). The sampled \latex sources are rendered and compared with the ground-truth images to judge their correctness, among which we collect 569,499 incorrect \latex sources and their corresponding ground-truth refinement. Fault localization and refinement models are fine-tuned independently, but both from the Nougat-base checkpoint and for one epoch, using a batch size of 32. The optimizer, and learning rate are the same as above.

\subsection{Table Models Training}
For table recognition, the generation model is fine-tuned on \tabletolatex (87,513 training, 5,000 validation, and 5,000 test instances). The other hyper-parameters are set in the same way as the fine-tuning of the generation model for formulae. The training data for fault localization and refinement models are collected in the same way as formulae, which contain 326,185 incorrect \latex sources and their ground-truth refinements. More details such as hyper-parameter tuning, and infrastructure are in supplementary materials.
\section{Results}
To evaluate \ours{} on \latex recognition, we study the following research questions:
\begin{itemize}
    \item \textbf{RQ1: What is the recognition accuracy of \ours{}?}
    \smallskip
    \item \textbf{RQ2: How is \ours{}'s iterative refinement ability?}
    \smallskip
    \item \textbf{RQ3: What is the impact of each design of \ours{}?}
\end{itemize}

Since \ours{} refines incorrect \latex sources iteratively before generating the correct ones or reaching the budget, we use \textsc{Latte$_k$} to present our approach with at most $k$ rounds of generation (one round of generation and $k-1$ round of refinements). \oursone refers to the result of only generating the initial draft using the \ours's generation model $M_G$. \ourstwo, \oursthree, and \oursfour refer to the result of letting \ours refine the incorrect \latex sources for one, two, and three rounds. We let \ours{} refine at most three rounds.

\subsection{RQ1: \ours Recognition Accuracy}
We used the five metrics for evaluation. \textbf{Match} (exact match accuracy) requires the rendered generated \latex source to have the same pixel values as the ground-truth image. \textbf{CW-SSIM}~\cite{cw-ssim} (complex-wavelet structural similarity index) measures the structural similarity between rendered and ground-truth images (we use MATLAB's implementation~\cite{matlab-cw-ssim} with \code{level=4, or=8, K=0.01}). \textbf{BLEU}~\cite{bleu} measures the textual similarity between the generated \latex source and the ground-truth \latex source (we report BLEU-4). \textbf{Edit} measures the column-wised edit distance between the rendered image and the ground-truth image, calculated by $1 - \frac{\text{column-wised edit distance}}{\text{number of pixel columns}}$. Lastly, we report the used time per sample for available techniques.

We compare \ours{} with a wide range of previous SOTAs~\cite{related5, related1, related4, related8, im2latex}, and also other MLLMs including a Vary-1.8B~\cite{vary} model fully fine-tuned using the training data, and a Llava-v1.5-7B~\cite{llava1.5} model fine-tuned using LoRA~\cite{lora}. Lastly, we also report the performance of commercial tools such as GPT-4V, Gemini-1.5-Pro, and Mathpix (commercial software for \latex recognition).

\begin{table}[!t]
    \scriptsize
    \centering
    \caption{Evaluation on \imgtolatex.}
    \begin{tabular}{l@{\hspace{8pt}}c@{\hspace{8pt}}c@{\hspace{8pt}}c@{\hspace{8pt}}c@{\hspace{8pt}}c}
         \toprule
         \textbf{Method} & \textbf{Match $\uparrow$}  & \textbf{CW-SSIM $\uparrow$} & \textbf{BLEU $\uparrow$} & \textbf{Edit $\uparrow$} & \textbf{Time $\downarrow$} \\
         \midrule
         WYGIWYS & 77.46 & - & 87.73 & 87.60 & - \\
         Double Attention & 79.81 & - & 88.42 & 88.57 & - \\
         EDPA & 82.07 & - & 92.31 & 91.39 & - \\
         WAP & 82.08 & - & 88.21 & 89.58 & - \\
         MI2LaTeX & 82.33 & - & 90.28 & 91.90 & - \\
         ConvMath & 83.41 & - & 88.33 & 90.80 & - \\
    \midrule
         Vary-1.8B & 11.91 & 0.7895 & 69.46 & 63.47 & 2.27s\\
         Llava-v1.5-7B & 13.54 & 0.7548 & 75.40 & 64.61 & 2.29s\\
    \midrule
         \oursone & 82.27 & 0.9462 & 92.91 & 93.11 & 0.87s \\
         \ourstwo & \textbf{90.44} & \textbf{0.9844} & \textbf{93.25} & \textbf{97.69} & 1.53s \\
         \bottomrule
    \end{tabular}
    \label{tab:main_result_formula}
\end{table}

\begin{table}[!t]
    \scriptsize
    \centering
    \caption{Evaluation on \tabletolatex.}
    \begin{tabular}{l@{\hspace{10pt}}c@{\hspace{10pt}}c@{\hspace{10pt}}c@{\hspace{10pt}}c@{\hspace{10pt}}c}
         \toprule
         \textbf{Method} & \textbf{Match $\uparrow$}  & \textbf{CW-SSIM $\uparrow$} & \textbf{BLEU $\uparrow$} & \textbf{Edit $\uparrow$} & \textbf{Time $\downarrow$} \\
         \midrule
         Vary-1.8B & 6.92 & 0.6253& 62.89 & 30.50 & 7.13s \\
         Llava-v1.5-7B & 13.90 & 0.7278 & 64.19 & 39.84 & 6.13s \\
    \midrule
         \oursone & 45.20 & 0.8128 & 79.06  & 73.82 & 2.24s\\
         \ourstwo & \textbf{59.18} & \textbf{0.8221} & \textbf{83.81}  & \textbf{77.51} & 5.34s \\
         \bottomrule
    \end{tabular}
    \label{tab:main_result_table}
\end{table}

\subsubsection{Formulae}
\Cref{tab:main_result_formula} shows the evaluation of \oursone and \ourstwo{} (the study of \oursthree and \oursfour are in RQ2). 
On \imgtolatex benchmark, with one round of refinement, \ourstwo{} successfully refines 823 incorrect \latex sources from \oursone and achieves 90.44\% Match, significantly outperforming all the existing state-of-the-art techniques. The CW-SSIM, BLEU, and Edit scores are also improved with the refinement by 0.0382, 0.34\%, and 4.58\%. Analysis of the significance of improvements can be found in the supplementary materials.

\subsubsection{Tables}
\Cref{tab:main_result_table} shows the evaluation results on \tabletolatex. As there are no open-sourced tools we can directly run on table recognition, we compare \oursone and \ourstwo with the fine-tuned Vary-1.8B and Llava-v1.5-7B. \oursone{}'s fine-tuned generation model generates \latex sources for tables matching 2,260 samples out of 5,000 (45.20\% Match). The lower scores of the evaluation metrics suggest the challenge and potential for improvement of \latex table recognition. With one round of refinement, \ourstwo correctly refines 699 incorrect sources and boosts the Match to 59.18\%. The CW-SSIM, BLEU, and Edit scores are also improved by 0.0093, 4.75\%, and 3.68\%. Both \oursone and \ourstwo significantly outperform the other MLLMs we fine-tuned.

\begin{table}[!t]
    \centering
    \scriptsize
    \setlength{\tabcolsep}{5pt}
    \caption{Comparison with commercial tools on 100 samples from \imgtolatex and \tabletolatex. The numbers are shown as ``\texttt{x}\textbar \texttt{y}'', where \texttt{x} is the result on \imgtolatex and \texttt{y} is the result on \tabletolatex.}
    \begin{tabular}{l|cccc}
    \toprule
         \textbf{Method} & \textbf{Match $\uparrow$}  & \textbf{CW-SSIM $\uparrow$} & \textbf{BLEU $\uparrow$} & \textbf{Edit $\uparrow$} \\
    \midrule
         GPT-4V$_1$ & \phantom{0}3.00 \textbar\ \phantom{0}2.00 & 0.7480 \textbar\ 0.5189 & 52.77 \textbar\ 49.56 & 61.25 \textbar\ 8.98 \\
         GPT-4V$_2$ & \phantom{0}7.00 \textbar\ \phantom{0}2.00 & 0.7212 \textbar\ 0.5059 & 50.87 \textbar\ 44.22 & 59.46 \textbar\ 5.64 \\
         Gemini$_1$ & 19.00 \textbar\ \phantom{0}0.00 & 0.6485 \textbar\ 0.3482 & 21.47 \textbar\ 35.19 & 63.60 \textbar\ 0.94 \\
         Gemini$_2$ & 19.00 \textbar\ \phantom{0}0.00 & 0.6191 \textbar\ 0.3911 & 25.78 \textbar\ 37.59 & 61.58 \textbar\ 1.27 \\
         Mathpix & 20.00 \textbar\ 11.00 & 0.8684 \textbar\ 0.6749 & 20.71 \textbar\ 49.45 & 84.44 \textbar\ 28.31 \\
    \midrule
         \oursone & 77.00 \textbar\ 40.00 & \textbf{0.9878} \textbar\ 0.8659 & 92.45 \textbar\ 77.53 & \textbf{97.68} \textbar\ 67.49 \\
         \ourstwo& \textbf{87.00} \textbar\ \textbf{67.00} & 0.9778 \textbar\ \textbf{0.8723} & \textbf{93.72} \textbar\ \textbf{83.82} & 96.92 \textbar\ \textbf{77.36} \\
    \bottomrule
    \end{tabular}
    \label{tab:comparison_commercial}
\end{table}

\begin{figure}[!t]
    \centering
    \includegraphics[width=\linewidth]{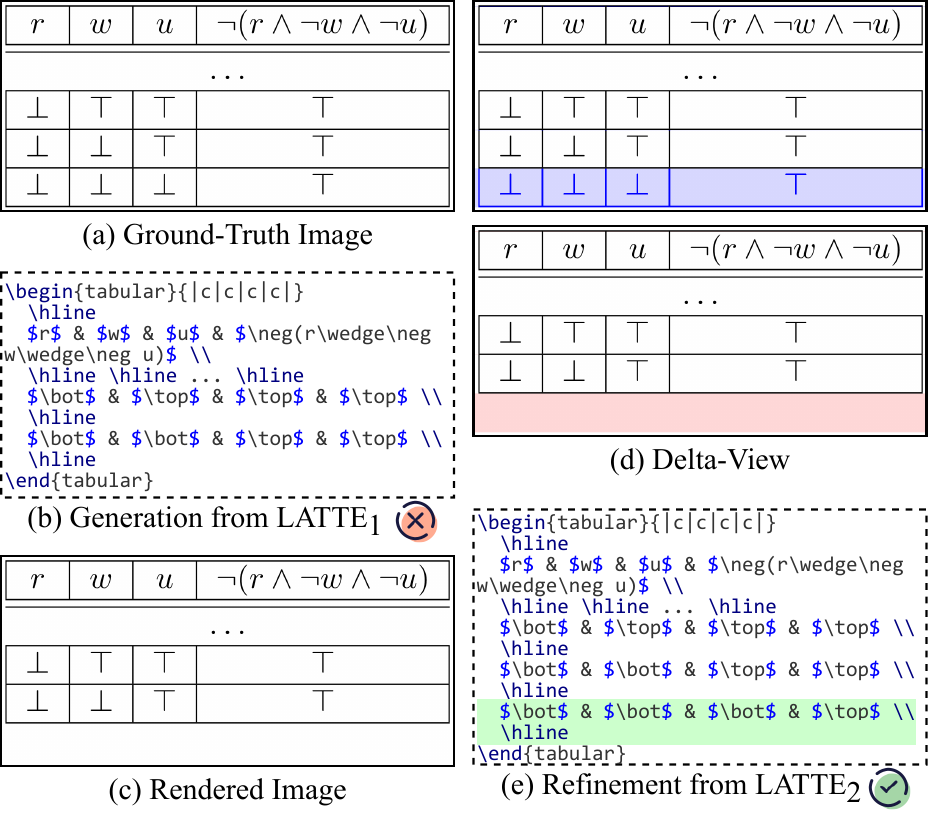}
    \caption{Example of \ourstwo{}'s correct refinement.}
    \label{fig:table_case_study}
\end{figure}

\subsubsection{Comparing with Commercial Tools}
\Cref{tab:comparison_commercial} shows the comparison with commercial tools on a subset of 100 samples from \imgtolatex and \tabletolatex.  Similarly, we use GPT-4V$_1$ to refer to prompting GPT-4V to generate the initial draft of \latex source, and GPT-4V$_2$ to refer to prompting it for one round of refinement of the incorrect sources (same for Gemini-1.5-Pro, and Mathpix is not applicable for refinement). We use few-shot learning~\cite{fsl} with three shots provided when prompting GPT-4V and Gemini-1.5-Pro for generation and refinement. Details can be found in the supplemental materials. 

\Cref{tab:comparison_commercial} shows that GPT-4V, Gemeni-1.5-Pro and Mathpix fail to generate the correct \latex source code most of the time. They also do not show the ability to effectively refine the incorrect \latex source with rendering feedback. Both \oursone and \ourstwo generate significantly more correct \latex sources than commercial MLLMs and software.

\subsubsection{Case Study}
\Cref{fig:table_case_study} shows an example, for which \oursone{} generates the incorrect source, missing the last row of the table (subfigures (b) and (c)). Yet, with the effective \deltaimage (subfigure (d)), \ourstwo{} successfully refines it and produces the correct source (subfigure (e)). For the same example, GPT-4V, Gemini-1.5-Pro, and Mathpix neither generate the correct source nor correctly refine it. More examples are provided in the supplementary materials.

\begin{figure}[!t]
    \centering
    \begin{minipage}{0.233\textwidth}
        \includegraphics[width=\linewidth]{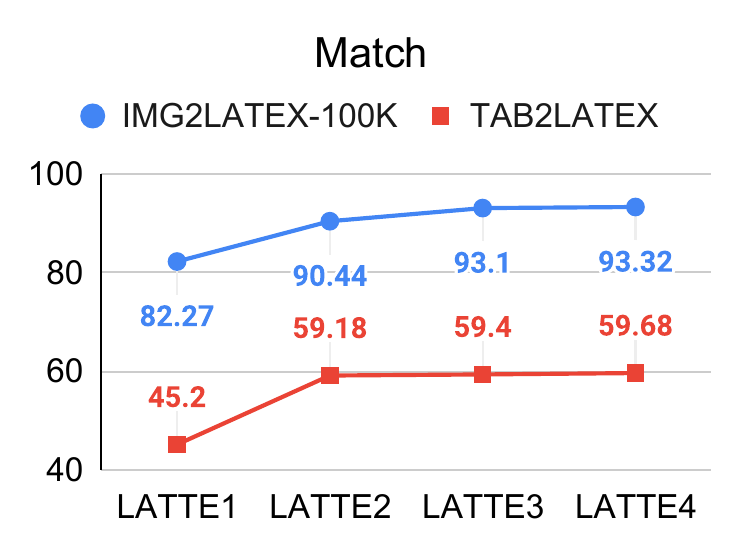}
    \end{minipage}
    \begin{minipage}{0.233\textwidth}
        \includegraphics[width=\linewidth]{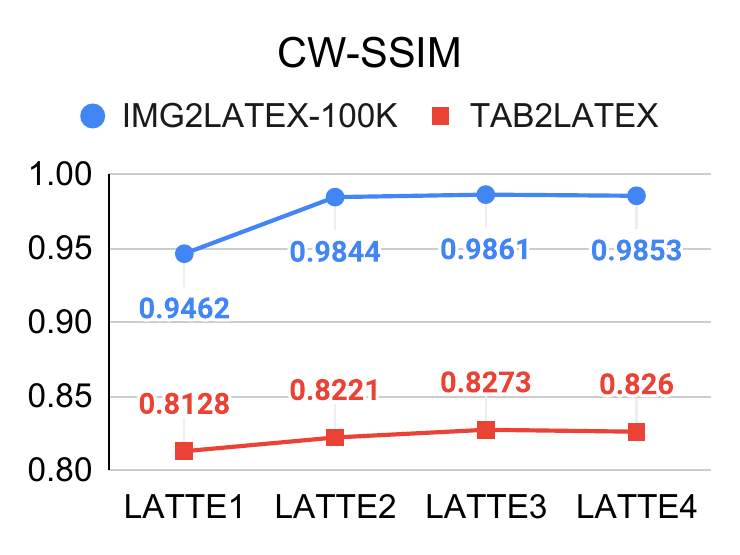}
    \end{minipage}
    \caption{Evaluation of \oursone to \oursfour.}
    \label{fig:iterative}
\end{figure}

\subsection{RQ2: \ours's Iterative Refinement Ability}
\Cref{fig:iterative} shows the results when \ours{} refines multiple rounds. On \imgtolatex, \ours{}'s Match keeps increasing, from 82.27\% to 93.32\% after three rounds of refinement, with the most significant improvement during the first refinement iteration. Similarly, for \tabletolatex, \ours{}'s Match increases from 45.20\% to 59.68\% in three rounds of refinement, and the biggest gain also happens at the first refinement round, by 13.99\%.  
For CW-SSIM, \ours's performance on \imgtolatex first increases from 0.9462 to 0.9844, which is a big gain of 0.0382, then fluctuates around it. The improvements on \tabletolatex are more moderate compared to that of \imgtolatex, by 0.0930 in the first round, with the remaining rounds staying around the same values.

Overall, \ours shows the ability to consistently improve the Match result by conducting iterative refinements, while the first round of refinement brings the most improvements.

\subsection{RQ3: Impact of Each Component of \ours}
\label{subsec:rq3}
\ours{} contains two designs: the \deltaimage feedback, and the fault localization model. To illustrate the effectiveness of each component, we design an ablation study by comparing \ours{} with the following variants (only one round of refinement is conducted using each method):
\begin{itemize}
    \item \ete is \ours without fault localization and \deltaimage. The refinement model generates a new \latex source (instead of starting from the fault location), given the original ground truth image.
    \item \etedv is \ours without fault localization. The refinement model generates a new \latex source, with  \deltaimage as the feedback.
\end{itemize}

\Cref{tab:ablation} shows the comparison between \ete, \etedv and \ours. Match and refinement rate (the number of correct refinements divided by the total number of incorrect sources that need refinement) are reported. By using \deltaimage as feedback, \etedv outperforms \ete on formulae by 2.13\% more Match (88.55\% vs. 86.42\%), and 12.04\% higher refinement rate (35.44\% vs. 23.40\%). On tables dataset, \etedv benefits from \deltaimage by improving the Match from 49.86\% to 59.52\% and refinement rate from 8.50\% to 26.13\%. Results show that \deltaimage is much more effective than just providing the ground-truth image.

As for the fault localization model, when comparing \ours{} and \etedv, the fault localization model helps the refinement model refine more incorrect formulae (46.08\% vs. 35.44\%), further increasing the Match from 88.55\% to 90.44\%. On the tables dataset, a slight decrease is observed, that using the fault localization model decreases the Match by 0.34\%. Such a decrease may be due to the lower fault localization accuracy on tables than on formulae.

\begin{table}[!t]
    \scriptsize
    \centering
    \setlength{\tabcolsep}{6pt}
    \caption{Impact of Fault Localization and \code{Delta-View}.}
    \begin{tabular}{lcc@{\hspace{5pt}}c@{\hspace{5pt}}cc}
         \toprule
         \multirow{2}{*}{\textbf{Method}} 
         & \multicolumn{2}{c}{\imgtolatex} & & \multicolumn{2}{c}{\tabletolatex} \\
         \cmidrule{2-3} \cmidrule{5-6}
         & \textbf{Match $\uparrow$} & \textbf{Refine. Rate $\uparrow$} &&  \textbf{Match $\uparrow$}  & \textbf{Refine. Rate $\uparrow$} \\
         \midrule
         \ete & 86.42 & 23.40 && 49.86 & 8.50 \\
         \etedv & 88.55 & 35.44 && \textbf{59.52} & \textbf{26.13} \\
         \ours & \textbf{90.44} & \textbf{46.08} && 59.18 & 25.51 \\
         \hline
         \cellcolor{lightgray}\ours{}$^*$ & \cellcolor{lightgray}90.45 & \cellcolor{lightgray}46.14 && \cellcolor{lightgray}59.72 & \cellcolor{lightgray}26.50\\
         \bottomrule
    \end{tabular}
    \label{tab:ablation}
\end{table}

\begin{figure}[!t]
    \centering
    \begin{minipage}{0.233\textwidth}
        \includegraphics[width=\linewidth]{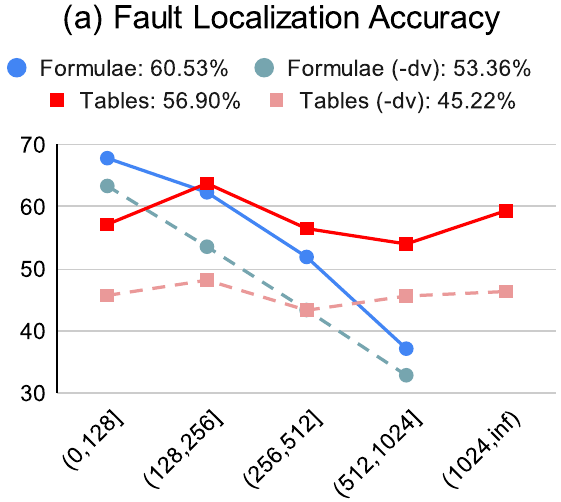}
    \end{minipage}
    \begin{minipage}{0.233\textwidth}
        \includegraphics[width=\linewidth]{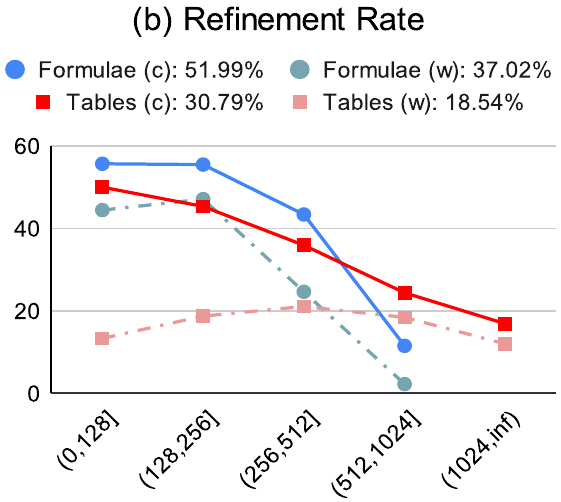}
    \end{minipage}
    \caption{(a) Fault localization accuracy with \deltaimage and with ground-truth image (-dv). (b) Success refinement rate under correct (c) or wrong (w) fault localization. The x-axis is the length, in the number of characters, of the incorrect \latex to be refined.}
    \label{fig:len}
\end{figure}

To better understand the impact of fault localization, \Cref{fig:len} (a) shows the fault localization accuracy on formulae and tables scripts with different lengths. On formulae, the fault localization accuracy (with \deltaimage or not) drops dramatically when the incorrect \latex scripts get longer. Yet the accuracy when using \deltaimage as feedback still consistently outperforms the accuracy when using the ground-truth image as feedback and the overall average is 60.53\% versus 53.36\%. Surprisingly, we find the fault localization accuracy does not drop much with longer incorrect tables, although the overall average accuracy of using \deltaimage is much higher (56.90\% versus 45.22\%).

\Cref{fig:len} (b) shows the refinement model's successful refinement rate on incorrect formulae and tables with different lengths. On formulae, the refinement rate also drops significantly when the incorrect \latex script gets longer. Additionally, when the fault localization model predicts the faulty location correctly, the refinement model has a higher success rate.
On the tables dataset, the trend is slightly different, as the refinement rate is always low if the fault localization model predicts incorrect faulty locations.

\subsubsection{Potential Improvement}
By digging into the fault localization accuracy, we see potential space for improvement by combining \ours{} with \etedv, referred to as \ours{}$^*$ in~\Cref{tab:ablation}. \ours{}$^*$ uses \etedv to refine incorrect \latex scripts with more than 512 characters, and uses \ours{} to refine those shorter than 512 characters. Such a simple ensemble reaches a higher Match and refinement rate than each single approach. 
Despite such potential, \ours{} is still the most effective single approach overall, and advanced ensemble approaches could be promising future work.
\section{Related Work}

\subsection{LaTeX Recognition}
Existing work on LR includes rule-based, grammar-based, and deep learning-based methods~\cite{related4}.
Learning-based solutions utilize the encoder-decoder architecture to tackle LR. The encoders often consist of convolution neural networks to extract image features with the decoders being recurrent neural networks to generate the output sequence in an end-to-end manner~\cite{related7,related2,related8,related1,related5,related9}. 
On table recognition, due to the difficulty, earlier work mainly focuses on table detection and table structure recognition~\cite{table2latex-related2-overview}. Recently, IBM researchers proposed an encoder-dual-decoder model to separately collect the structural information and contents within table cells~\cite{table2latex-related3-xml}. However, even with all such information, users still struggle with reproducing the renderable source of \latex tables, as the structure and content extracted cannot be combined. Conversely, another work proposes a dataset~\cite{table2latex-related1-report} containing pairs of table images and the corresponding \latex source code. This is the only work we have found that works on the end-to-end \latex table recognition, yet their dataset is not accessible anymore.

This work not only adds the iterative-refinement pipeline to the generation process but also includes a \faultloc to predict the faulty location for those incorrect sources to improve the recognition accuracy. The proposed \tabletolatex dataset contains 97K well-filtered table images and source code pairs and fills the blank of the end-to-end \latex tables recognition dataset.

\subsection{Iterative Refinement Framework}
Researchers have identified the process of refining one's answer as a typical part of the problem-solving process~\cite{selfrefine-related2,selfrefine-related1}, which has been added to numerous fields, \eg, program repair, code, and text generation. Existing work on the program repair applies automatic refinement for repair and fault localization on imperative programs based on symbolic execution~\cite{faultloc-related1}. For code and text generation, some work prompts the LLM to provide feedback by itself, then refine its answer without additional training~\cite{llm-selfrefine-related1, llm-selfrefine-related4}, while others fine-tune the LLMs to enhance their ability to adopt the feedback~\cite{llm-selfrefine-related2, llm-selfrefine-related3} for better refinement.
This work is the first to apply the iterative refinement framework within the field of LR. Applying iterative-refinement in LR is new and has unique challenges regarding providing effective image feedback. \ours introduces \deltaimage as novel feedback to address such challenges in multi-modal generation and refinement, which is shown to be helpful in the ablation study.

\subsection{Multi-Modal Large Language Models}
Many MLLMs have shown great ability in vision and text tasks, such as image captioning~\cite{clip, blip, blip2}, image understanding~\cite{pix2struct}, visual question answering~\cite{blip,blip2,llava,llava1.5, instructblip}, etc. The early paradigm for building MLLMs involves jointly training vision and text models, \eg, CLIP and BLIP~\cite{clip, blip}. Later work train adapters to connect the pre-trained vision encoder and text decoder, to borrow the strong text generation ability of textual LLMs without retraining from scratch~\cite{llava, instructblip, llava1.5, blip2, Qwen-VL, sharegpt4v}. 
However, most existing MLLMs (including GPT-4V) are optimized for understanding pictures and natural language, not document images and \latex. Nougat~\cite{nougat} is the only existing MLLM pre-trained on documents, on which we build \ours{}'s models.

\section{Limitation}
\noindent One limitation of our work is that we only explore Nougat as \ours{'s} backbone model. Many other MLLMs, such as Llava, can be used as our backbone model. However, they are mostly pre-trained on natural language and pictures and show very poor performance on \latex{} recognition. Nougat is the best MLLM we can find that is dominantly pre-trained on documents.
Another possible limitation of \ours is that existing metrics evaluate \ours and other related \latex recognition work cannot reflect human's preference on rendered \latex formulae or tables results. These metrics are either very harsh, i.e., pixel level matching, or only consider one-dimensional column-wise matching. To make our evaluation more robust, we add CW-SSIM to investigate the structural similarity of the rendered formulae or tables, but there could be potentially better metrics.

\section{Conclusion}
This work proposes \ours{}, the first iterative-refinement approach for Latex recognition for both formulae and tables. \ours{} uses a generation model to produce \latex sources from images; and builds a fault localization model and a refinement model to refine the generated \latex source iteratively. To provide effective feedback to the iterative process, this work proposes the \idiff algorithm, which generates \deltaimage that pinpoints the difference between the ground truth and rendered images. This work also constructs a new \latex table recognition dataset \tabletolatex. 
With one round of refinement, \ours{} outperforms existing techniques by 7.03\% of the exact match on \latex formulae recognition. Besides, \ours{}'s formulae and table recognition ability exceed commercial tools by a significant margin, showing great generalizability and effectiveness. 
In the future, it would be promising to develop better algorithms for pinpointing image differences for tables and formulae to boost the performance of our iterative refinement approach.

\section*{ACKNOWLEDGMENTS}
This research was supported in part by NSF 1901242 and 2006688 and a CFI fund.

\bibliography{aaai25}

\section*{Appendix}

\subsection{\texttt{ImageEdit} Algorithm}
\label{sec:algorithm}
\Cref{alg:idiff} shows the \idiff algorithm for generating the column-wised \deltaimage, whose input is $I_g$ (the expected ground-truth image) and $I_r$ (the image rendered from the generated incorrect \latex source). \idiff first uses the \code{Wagner$-$Fischer$^*$} (\cref{alg:edit-distance}) to calculate the least number of insertions, deletions, and substitutions of columns needed to transform the rendered image to the ground truth image. These insertion, deletion, and substitution operations are represented in the \deltaimage using the \code{ShowDiff} algorithm.

\RestyleAlgo{ruled}
\begin{algorithm}
\footnotesize
\caption{\idiff \code{(Column-Wised)}}
\label{alg:idiff}

\SetKwInput{KwInput}{Input}
\SetKwInput{KwOutput}{Output}
\KwInput{$I_g, I_r \in \mathbb{N}^{H\times W\times 3}$, the ground-truth image and rendered image. Images are viewed as a list of $W$ number of pixel columns ($\mathbb{N}^{H\times 3}$).}
\KwOutput{$\Delta(I_g, I_r)$, the delta-view feedback.}
\DontPrintSemicolon

$I_W \gets [[255,255,255]\times H]$ \\
$ops \gets$ \texttt{Wagner$-$Fischer$^*$}\text{($I_g, I_r$)} \\
\For{$(op, i, j)$ \text{\textbf{in}} $ops$}{
        \If{$op = \textit{``Delete"}$}{
            $I_r[i] \gets$ \texttt{ShowDiff}$(I_W, I_r[i])[1]$\;
        }\ElseIf{$op = \textit{``Insert"}$}{
            $I_g[j] \gets$ \texttt{ShowDiff}$(I_g[j], I_W)[0]$\;
        }\Else{
            $I_g[j], I_r[i] \gets$ \texttt{ShowDiff}$(I_g[j], I_r[i]$)\;
        }
}
\textbf{Return} {$I,I_i$}\;
\end{algorithm}
\begin{algorithm}
\footnotesize
\caption{\code{ShowDiff}}
\label{alg:highlight}
\textbf{Input: }{$I_g, I_r$}

$I_g[I_g = [255,255,255]] \gets [255, 200, 200]$\\
$I_r[I_r = [255,255,255]] \gets [200, 200, 255]$\\
$I_g[(I_g \neq I_r)\And(I_r = I_W)] \gets [255, 0, 0]$\\
$I_r[(I_g \neq I_r)\And(I_g = I_W)] \gets [0, 0, 255]$\\
\textbf{Return} $I_g, I_r$
\end{algorithm}

The original Wagner–Fischer~\cite{related-LDdistance-algo} algorithm is a dynamic programming algorithm that computes the edit distance between two strings of characters. We modify it to work for two images, by treating an image as a list of column pixels, referred to as the \code{Wagner$-$Fischer$^*$} algorithm (\cref{alg:edit-distance}). In addition, our \code{Wagner$-$Fischer$^*$} algorithm backtracks to find the needed operations (insertions, deletions, and substitutions) to transform the rendered image to the ground truth image.

\begin{algorithm}
\footnotesize
\caption{\code{Wagner$–$Fischer$^*$}}
\label{alg:edit-distance}

\SetKwInput{KwInput}{Input}
\SetKwInput{KwOutput}{Output}
\KwInput{$I_g, I_r \in \mathbb{N}^{H\times W\times 3}$, the ground-truth image and rendered image. Images are viewed as a list of $W$ number of pixel columns ($\mathbb{N}^{H\times 3}$).}
\KwOutput{$ops$, operations to be performed on $I_g$ and $I_r$.}
\DontPrintSemicolon

\tcp{Dynamic programming to calculate the Levenshtein-Distance}
$D \gets [[0, \textit{``"}] \times (W+1)] \times (W+1)$\;
$D[i \gets 0, \ldots, W+1][0] \gets [I, \textit{``Delete"}]$\;
$D[0][j \gets 0, \ldots, W+1] \gets [j, \textit{``Insert"}]$\;

\For{$i \gets 1,\ \ldots,W+1$}{
    \For{$j \gets 1,\ \ldots,W+1$}{
        \eIf{$I_g[i-1] = I_r[j-1]$}{
            $Cost_{sub} \gets D[i-1][j-1][0]$\;
            $same \gets \bf{true}$
        }{
            $Cost_{sub} \gets D[i-1][j-1][0] + 1$\;
            $same \gets \bf{false}$
        }
        $Cost_{del} \gets D[i-1][j][0] + 1$\;
        $Cost_{ins} \gets D[i][j-1][0] + 1$\;
        $Cost_{min} \gets \texttt{min}(Cost_{sub},\ Cost_{del},\ Cost_{ins})$\;
        \If{$Cost_{ins} = Cost_{min}$}{
            $D[i][j] \gets [Cost_{ins}, \textit{``Insert"}]$\;
        }\ElseIf {$Cost_{sub} = Cost_{min}$}{
            \eIf{same}{
                $D[i][j] \gets [Cost_{sub}, \textit{``Copy"}]$
            }{
                $D[i][j] \gets [Cost_{sub}, \textit{``Substitute"}]$
            }
        }\Else{
            $D[i][j] \gets [Cost_{del}, \textit{``Delete"}]$\;
        }
    }
}
\tcp{Backtrack to find the optimal operations sequence}
$ops \gets []$\;
$i, j \gets W, W$\;
\While{i $>$ 0 and j $>$ 0}{
    $op \gets D[i][j][1]$\;
    \If{$op = \textit{``Copy"}$}{
        $i,\ j \gets i - 1,\ j - 1$\;
    }\Else{
        \If{$op = \textit{``Delete"}$}{
            $ops.\texttt{append}(\textit{``Delete"},\ i-1,\ \textbf{null})$\;
            $i \gets i - 1$\;
        }\ElseIf{$op = \textit{``Insert"}$}{
            $ops.\texttt{append}(\textit{``Insert"},\ \textbf{null},\ j - 1)$\;
            $j \gets j - 1$\;
        }\Else{$op = \textit{``Substitute"}$\;
            $ops.\texttt{append}(\textit{``Substitute"},\ i - 1,\ j - 1)$\;
            $i,\ j \gets i - 1,\ j - 1$\;
        }
    }
}
\While{i $>$ 0}{
    $ops.\texttt{append}(\textit{``Delete"},\ i - 1,\ \textbf{null})$\;
    $i \gets i - 1$\;
}
\While{j $>$ 0}{
    $ops.\texttt{append}(\textit{``Insert"},\ \textbf{null},\ j - 1)$\;
    $j \gets j - 1$\;
}
\textbf{Return} \texttt{reversed}($ops$)
\end{algorithm}

\subsection{T{\small AB}2L{\small ATEX} Dataset}
For table recognition, as there are no open-sourced datasets for end-to-end \latex table recognition yet, this work constructs a new dataset, \textbf{\tabletolatex}. 

\tabletolatex consists of 97,532 rendered images of tables and their \latex sources. The \latex sources are collected from academic papers within these six distinct sub-fields of computer science---Artificial Intelligence, Computation and Language, Computer Vision and Pattern Recognition, Cryptography and Security, Programming Languages, and Software Engineering---from the arXiv repository, covering the years 2018 to 2023. Once the paper sources are downloaded, tables are identified and extracted from the \latex source code by matching \code{\textbackslash begin\{tabular\}} and \code{\textbackslash end\{tabular\}} and removing the comments. Then, the \latex table source scripts are rendered to PDF format and converted to PNG format at 160 dpi. In the final step, the rendered images are resized or padded to the resolution of $1344\times672$ pixels. 

When we are rendering the \latex sources, we enclose the table sources with the following context so that the \code{pdflatex} renderer can produce the PDF files.

\subsection{Experimental Setup Details}
In this section, we provide the setup of training each model, as well as the infrastructure we use.

\subsubsection{Formulae: Generation Model}
We use the default split of the \imgtolatex~\cite{im2latex} dataset, which has 73,812 training, 18,672 validation, and 10,072 test instances, to train the generation model. We tune the training epoch in the range of \code{[1,2]}, and learning rate in the range of \code{[$2e^{-5}$,$3e^{-5}$,$5e^{-5}$]}. We use random search to fine-tune four models and take the one with the highest Match on the validation set.
Eventually, we fine-tune the Nougat-base model~\cite{nougat} for two epochs, using a batch size of 16. The model weights are optimized using the AdamW~\cite{adamw} optimizer, with the learning rate set to $3e^{-5}$, using 1,000 steps of warm-up and adjusted by a cosine decay scheduler.

\subsubsection{Formulae: Fault Localization Model}
For fault localization models, we collect incorrect \latex sources by sampling 20 \latex sources per image in the training set of \imgtolatex (sampling temperature set to 0.8, max new tokens set to 512). The sampled \latex sources are rendered and compared with the ground-truth images to judge their correctness, among which we collect 569,499 incorrect \latex sources and their corresponding ground-truth refinement. The input is the incorrect \latex source, and the label is the index of the first token that is different from the ground-truth \latex source.

The fault localization model is fine-tuned from the Nougat-base checkpoint. We tune the training epoch in the range of \code{[1,2]}, and learning rate in the range of \code{[$2e^{-5}$,$3e^{-5}$,$5e^{-5}$]}. We use random search to fine-tune four models and take the one with the highest fault localization accuracy on the validation set. Finally, the fault localization model is fine-tuned for one epoch, using a batch size of 32. The optimizer is AdamW and the learning rate is $3e^{-5}$.

\subsubsection{Formulae: Refinement Model}
The refinement model uses the same training data as the fault localization model. The input is the incorrect \latex source constructed into the format as Figure 4 in the paper shows (using the ground-truth fault localization), and the output is the expected part from the correct \latex source to replace the faulty part. The refinement model is also fine-tuned from the Nougat-base checkpoint. Due to the computing cost, we do not tune the hyper-parameter and reuse the setting of fine-tuning the fault localization model to fine-tune the refinement model.

\subsubsection{Tables: Generation Model}
For table recognition, the generation model is fine-tuned on \tabletolatex (87,513 training, 5,000 validation, and 5,000 test instances). Similarly, we tune the training epoch in the range of \code{[1,2]}, and learning rate in the range of \code{[$2e^{-5}$,$3e^{-5}$,$5e^{-5}$]}. We use random search to fine-tune four models and take the one with the highest Match on the validation set.
Eventually, the generation model is also fine-tuned for two epochs (starting from the Nougat-base checkpoint), using a batch size of 16. The optimizer is AdamW, and the learning rate is $3e^{-5}$.

\subsubsection{Tables: Fault Localization Model}
The training data for fault localization and refinement models are collected in the same way as formulae, which contain 326,185 incorrect \latex sources and their ground truth. The hyper-parameters are tuned in the same way as that of formulae, and the final model is trained for one epoch, using a batch size of 16. The optimizer is AdamW and the learning rate is $3e^{-5}$.

\subsubsection{Tables: Refinement Model}
The refinement model uses the same training data as the fault localization model. We also reuse the hyper-parameters of fine-tuning the tables fault localization model to fine-tune the refinement model.

\subsubsection{Infrastructure}
\ours{}'s models are implemented with PyTorch and HuggingFace's Transformers, and the training script uses DeepSpeed. All the experiments are conducted on a cluster with 96 CPU cores, and four NVIDIA RTX A5000 GPUs (each with 24 GB memory).

\subsection{Prompting GPT-4V and Gemini-1.5-Pro}
In this section, we describe the prompts we use to query GPT-4V and Gemini-1.5-Pro for \latex source generation and refinement.

\subsubsection{Prompt for Generation}
\begin{figure}[!t]
    \centering
    \includegraphics[width=\linewidth]{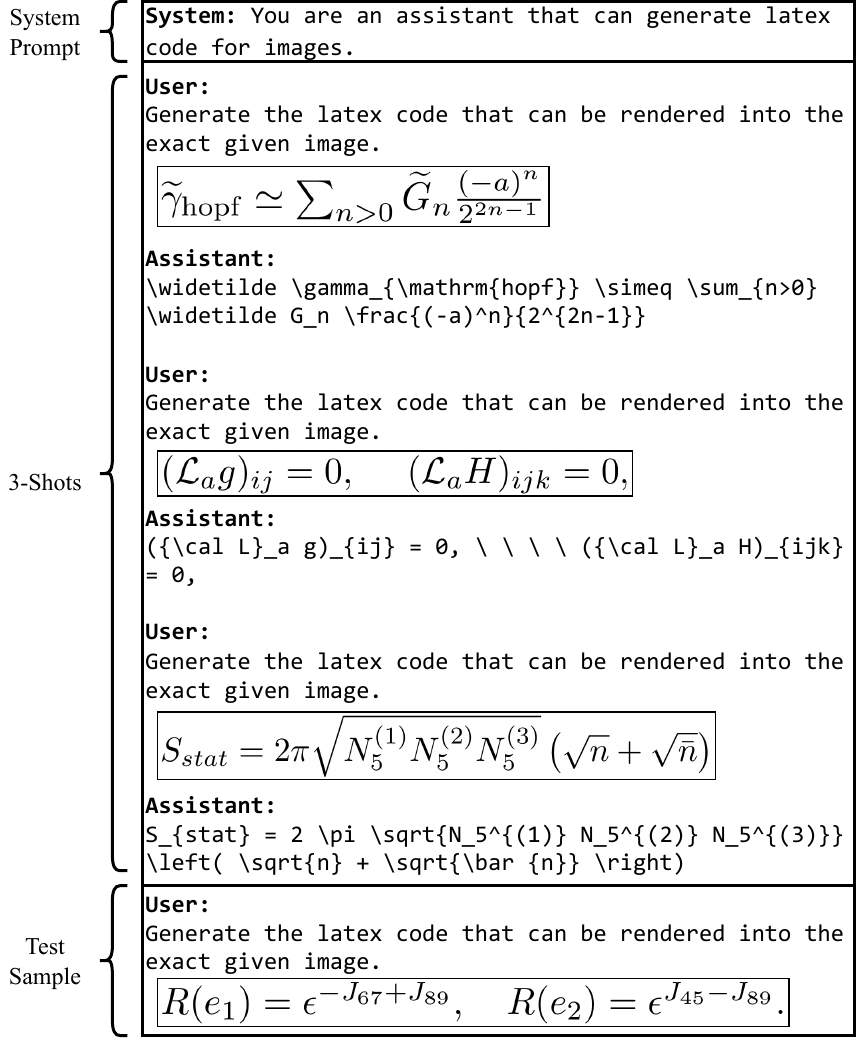}
    \caption{Prompt for Querying GPT-4V and Gemini-1.5-Pro for \latex Source Generation.}
    \label{fig:gpt4-generation}
\end{figure}

\Cref{fig:gpt4-generation} shows the prompt that we use for asking GPT-4V to generate the \latex source of a given image. To help GPT-4V better understand the task, besides the system prompt, we adopt few-shot learning and provide three examples. Each example contains a formula image and a sentence ``\code{Generate the latex code that can render into the exact given image}'' as the user's input, and the ground-truth \latex source as the expected model's output. Following the three examples, the image of the test sample is given, and GPT-4V is supposed to generate the \latex source of the test image following the format of the few-shot examples. 

\subsubsection{Prompt for Refinement}
\begin{figure}[!t]
    \centering
    \includegraphics[width=\linewidth]{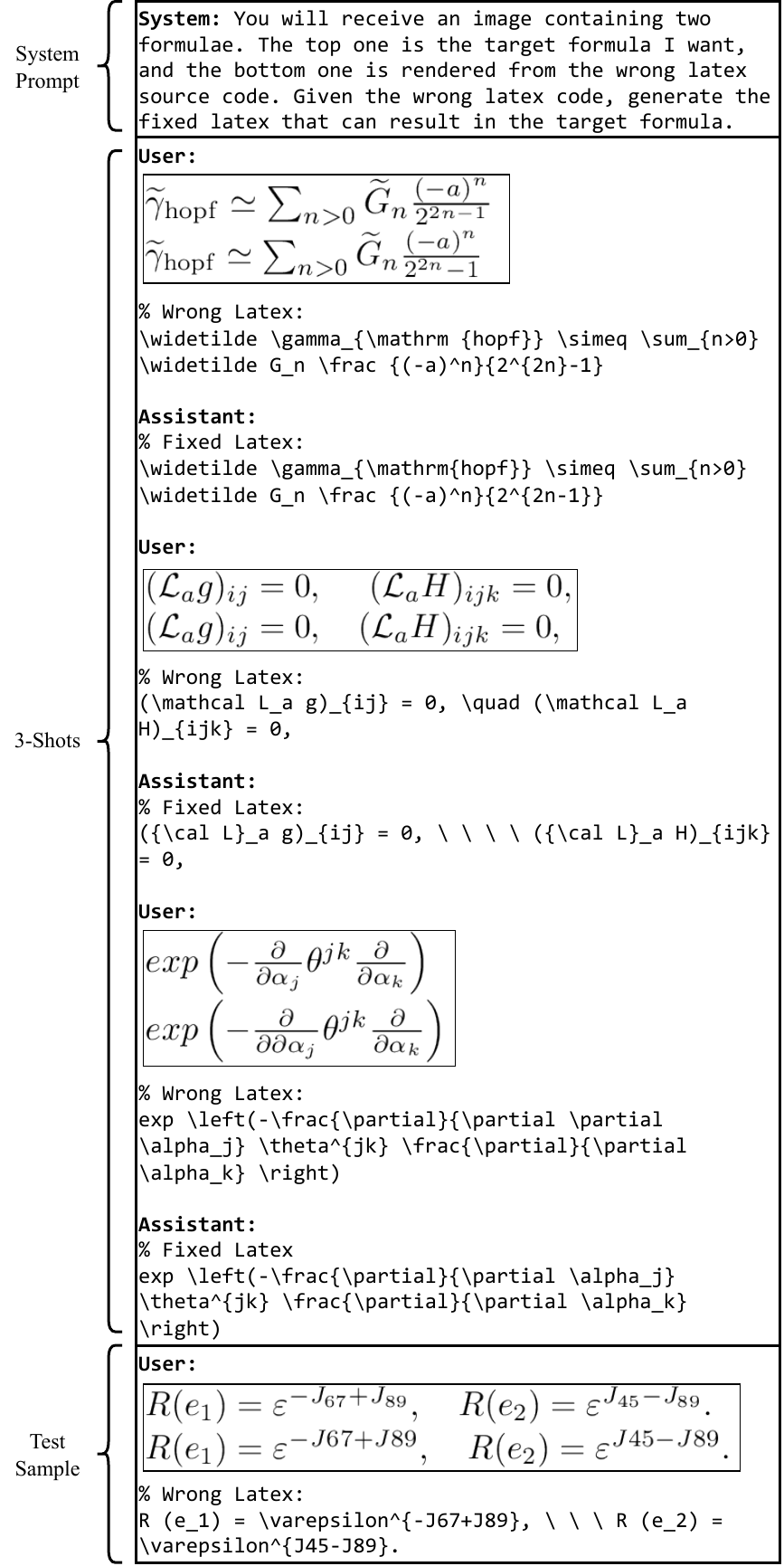}
    \caption{Prompt for Querying GPT-4V and Gemini-1.5-Pro for \latex Source Refinement.}
    \label{fig:gpt4-refinement}
\end{figure}

If GPT-4V generates the incorrect \latex source, we prompt it for refinement to also compare its refinement ability with \ours{}.~\Cref{fig:gpt4-refinement} shows the prompt and the three examples provided for few-shot learning. 

\begin{figure}[!t]
    \centering
    \includegraphics[width=\linewidth]{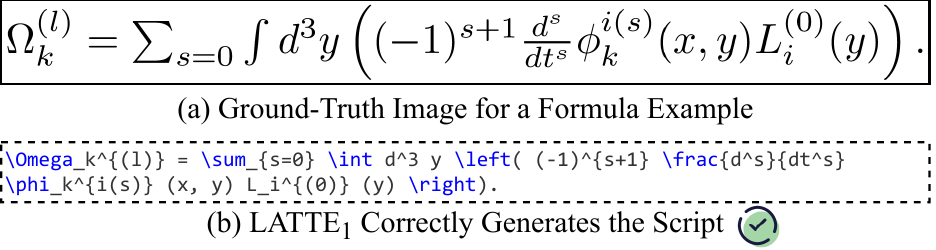}
    \caption{A Formula Example for Which \oursone Correctly Generates the \latex Source.}
    \label{fig:formula_correct_generation}
\end{figure}

\begin{figure}[!t]
    \centering
    \includegraphics[width=\linewidth]{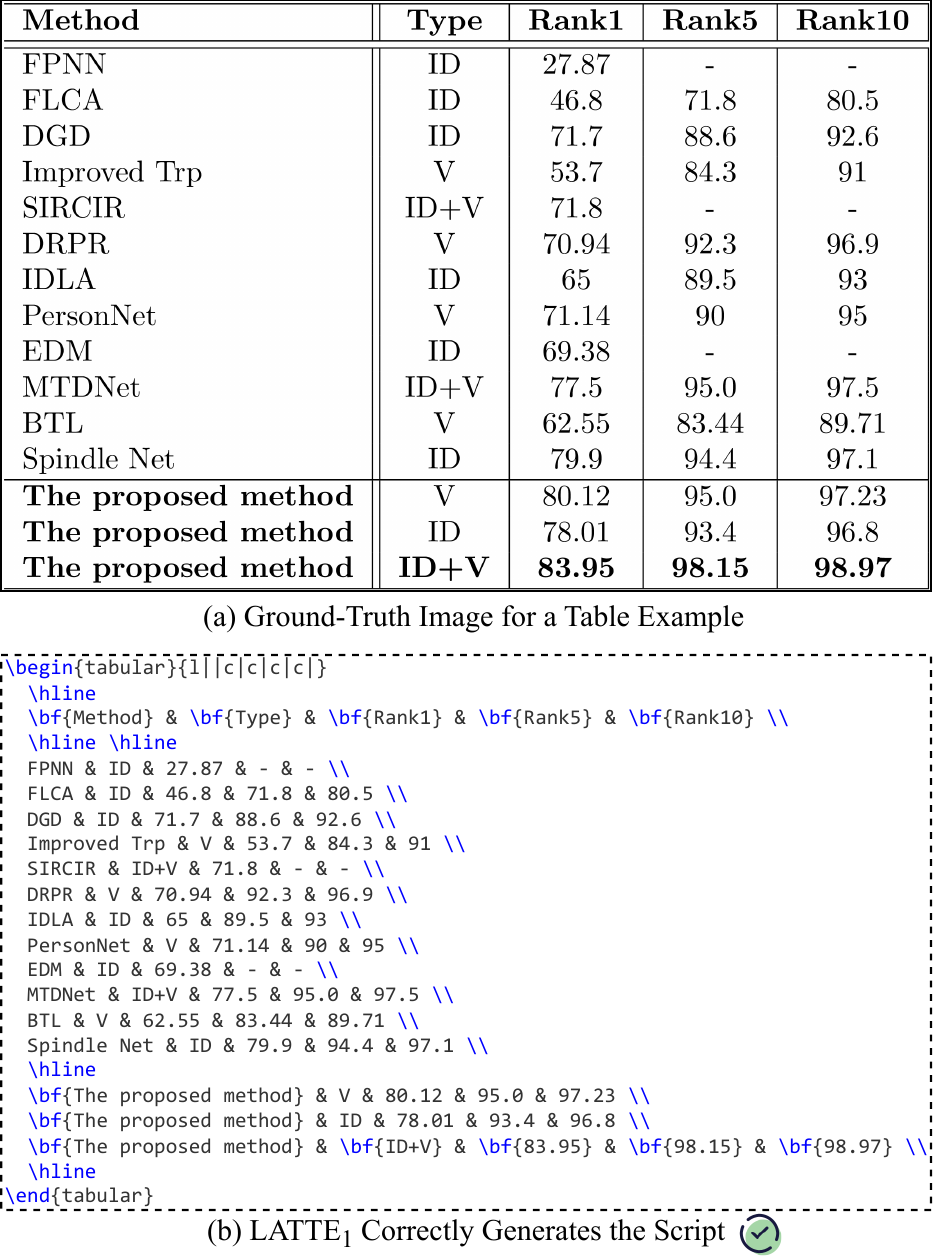}
    \caption{A Table Example for Which \oursone Correctly Generates the \latex Source.}
    \label{fig:table_correct_generation}
\end{figure}

\subsection{Additional Case Studies}

\subsubsection{Correction Recognition by Generation Model}
\ours{} can directly generate correct \latex source for 82.27\% formulae in the \imgtolatex benchmark and 45.20\% tables in the \tabletolatex benchmark without any refinement required (referred to as \oursone).

\Cref{fig:formula_correct_generation} and \Cref{fig:table_correct_generation} show two complex examples for which \oursone{} (the generation model only) correctly generates the sources in the first round, showing its strong \latex recognition capability.

\begin{figure}[!t]
    \centering
    \includegraphics[width=\linewidth]{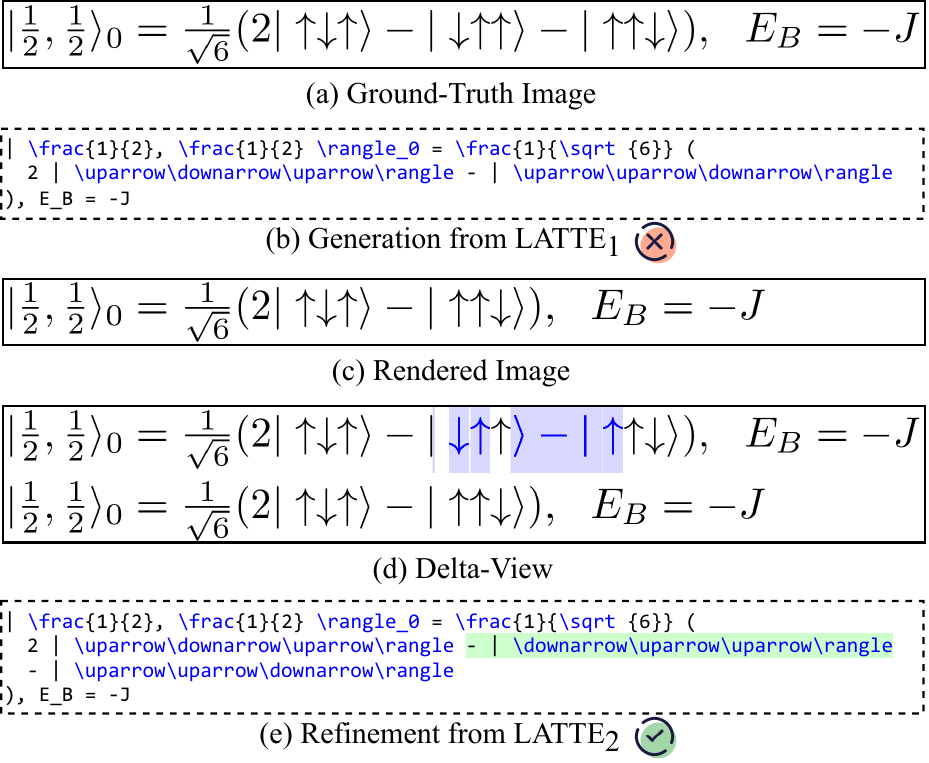}
    \caption{A Formula Example for which \ourstwo Correctly Refines the Incorrect \latex Source Generated by \oursone{}.}
    \label{fig:formula_correct_refinement}
\end{figure}

\begin{figure*}[htp]
    \centering
    \includegraphics[width=0.85\linewidth]{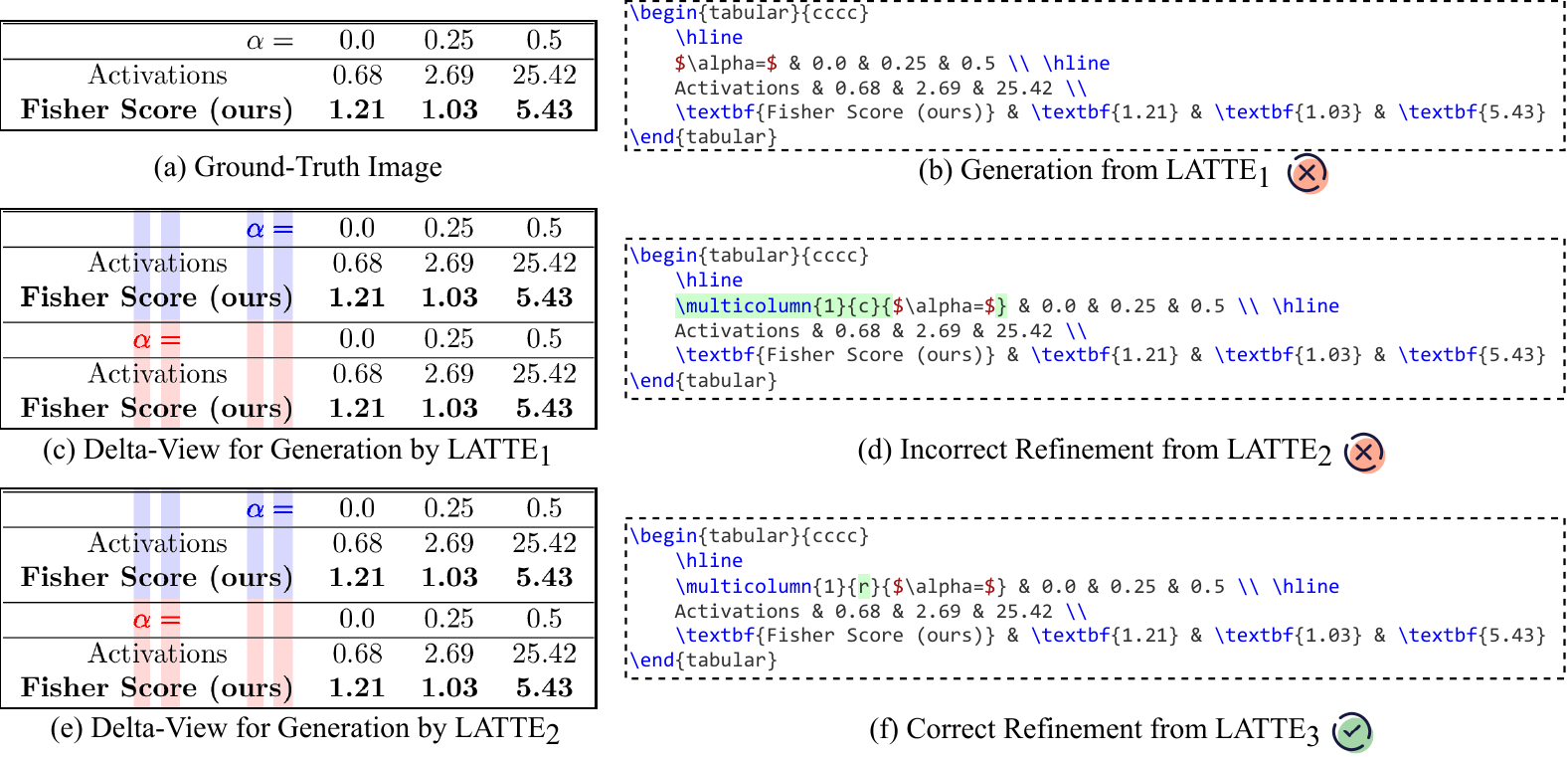}
    \caption{Example for Which Multiple Rounds of Refinements are Needed.}
    \label{fig:correct_iterative}
\end{figure*}

\subsubsection{Correct Refinement in One Round}
\Cref{fig:formula_correct_refinement} shows a formula example for which \oursone{} initially generates the incorrect \latex source. The incorrect source misses an operand, $-|\downarrow \uparrow \uparrow \rangle$, in the formula, which is correctly fixed by \ourstwo in one round of refinement.

\subsubsection{Correct Refinement in Multiple Rounds}
\Cref{fig:correct_iterative} shows an example for which multiple rounds of refinements are performed by \ours{} to obtain the final correct \latex source. \Cref{fig:correct_iterative} (a) shows the expected image, which contains a 3 $\times$ 4 table. The layout of the leftmost column is the key challenge in this example, where ``$\alpha =$'' is aligned to the right but ``{\small Activations}'' and ``\textbf{\small Fisher Score (ours)}'' are aligned to the center.

\oursone{} generates the initial \latex source as shown in \Cref{fig:correct_iterative} (b), where it simply uses center alignment for all the columns. The layout mismatch is highlighted in the \deltaimage in \Cref{fig:correct_iterative} (c), where ``$\alpha =$'' is expected to be aligned to the right, but the rendered image puts it in the center.

Given the \deltaimage in \Cref{fig:correct_iterative} (c) and the incorrect source in \Cref{fig:correct_iterative} (b), \ourstwo{} refines the \latex source as shown in \Cref{fig:correct_iterative} (d). The refinement encloses the expression ``$\alpha =$'' with the \code{\textbackslash multicolumn} environment. However, in the \code{\textbackslash multicolumn} environment, the content is still center-aligned. Thus, the refinement generated by \ourstwo{} is still incorrect and \deltaimage is generated as shown in \Cref{fig:correct_iterative} (e). 

Although incorrect, the failed refinement (\Cref{fig:correct_iterative} (d)) is \textbf{one step closer} to the ground truth. In the second try of refinement, given the \deltaimage and the failed refinement, \oursthree{} correctly fixes the \latex source by changing the \code{\textbackslash multicolumn} environment to be right-aligned.

\end{document}